\documentclass[10pt,twocolumn,letterpaper]{article}

\PassOptionsToPackage{table}{xcolor}
\usepackage[pagenumbers]{cvpr} 

\usepackage{multirow}
\usepackage{array}
\usepackage{tabularx}
\usepackage{colortbl}

\usepackage{pifont}

\usepackage{fontawesome5}

\definecolor{medalgold}{RGB}{212,175,55}
\definecolor{medalsilver}{RGB}{150,150,150}
\definecolor{medalbronze}{RGB}{205,127,50}

\usepackage{tcolorbox}

\newtcolorbox{findingtitle}{
  colback=gray!10,
  colframe=gray!40,
  boxrule=0.4pt,
  arc=1pt,
  left=5pt,
  right=5pt,
  top=3pt,
  bottom=3pt,
  before skip=5pt,
  after skip=3pt
}

\usepackage{float}
\usepackage{algorithm}
\usepackage{algorithmic}
\usepackage[disable]{todonotes}
\usepackage[T1]{fontenc}
\usepackage{booktabs}
\usepackage[font=small,labelfont=bf,labelsep=period]{caption}
\usepackage{booktabs,tabularx,array}
\definecolor{seamsection}{HTML}{F0F5FA}
\definecolor{seamsubgroup}{HTML}{F6F7F8}
\definecolor{cvprgray}{HTML}{F2F4F7}
\definecolor{avgblue}{HTML}{EBF2FA}

\newcommand{\skillupd}[1]{#1\rlap{\kern0.5pt\textsuperscript{\ensuremath{\dagger}}}}

\newcolumntype{R}[1]{>{\raggedleft\arraybackslash}p{#1}}

\usepackage{listings}
\tcbuselibrary{listings,breakable,skins}

\floatstyle{ruled}
\newfloat{listing}{tb}{lst}
\floatname{listing}{Listing}

\definecolor{cvprblue}{rgb}{0.21,0.49,0.74}
\usepackage[pagebackref,breaklinks,colorlinks,allcolors=cvprblue]{hyperref}

\def\confName{CVPR}
\def\confYear{2027}

\title{SEAM: A Self-Evolving and Adaptive Probing Framework for Automatic MLLM Hallucination Evaluation}

\author{Shiyu Zhang$^{1}$\thanks{Equal contribution.}, Zengyi Wo$^{1}$\footnotemark[1], Yuming Liao$^{1}$, Ran He$^{2}$, Caifeng Shan$^{1}$, Yanbo Fan$^{1}$\thanks{Corresponding author.}\\
{$^1$Nanjing University}\ \ {$^2$CASIA}\\
}

\begin{document}
\maketitle
\begin{abstract}
The severe susceptibility of multimodal large language models (MLLMs) to hallucinations poses a critical threat to their reliability, making a rigorous evaluation framework indispensable.
However, existing evaluation pipelines largely rely on predefined test items, suffering from high construction costs, rapid obsolescence as MLLMs evolve, and ineffectiveness in probing model-specific weaknesses.
To address these limitations, we propose \textbf{SEAM}, a self-evolving and adaptive probing framework for automatic MLLM hallucination evaluation. 
Specifically, SEAM introduces a composable Hallucination Probe Construction Library comprising ten hallucination-probing skills, three elicitation wrappers, and five question formats. Building on this library, SEAM establishes a dual-loop framework. In the inner loop (adaptive probing), skill-guided visual analysis, question composition, and hierarchical quality verification are coordinated with verifier and target-response feedback to refine test items and uncover model-specific hallucinations. In the outer loop (skill evolution), a skill updater distills historical probing trajectories into localized, reusable rule updates, which undergo held-out validation and are frozen before testing. 
Driven by these innovative structural designs, SEAM establishes an automated and self-evolving evaluation paradigm.
Experiments across four target MLLMs and ten hallucination types demonstrate that SEAM substantially outperforms representative static benchmarks, achieving an 89.2\% relative improvement in mean valid hallucination discovery rate across 12 matched comparisons.
\end{abstract}
    
\section{Introduction}
\label{sec:intro}

\begin{figure}[t]
    \centering
    \includegraphics[width=\columnwidth]{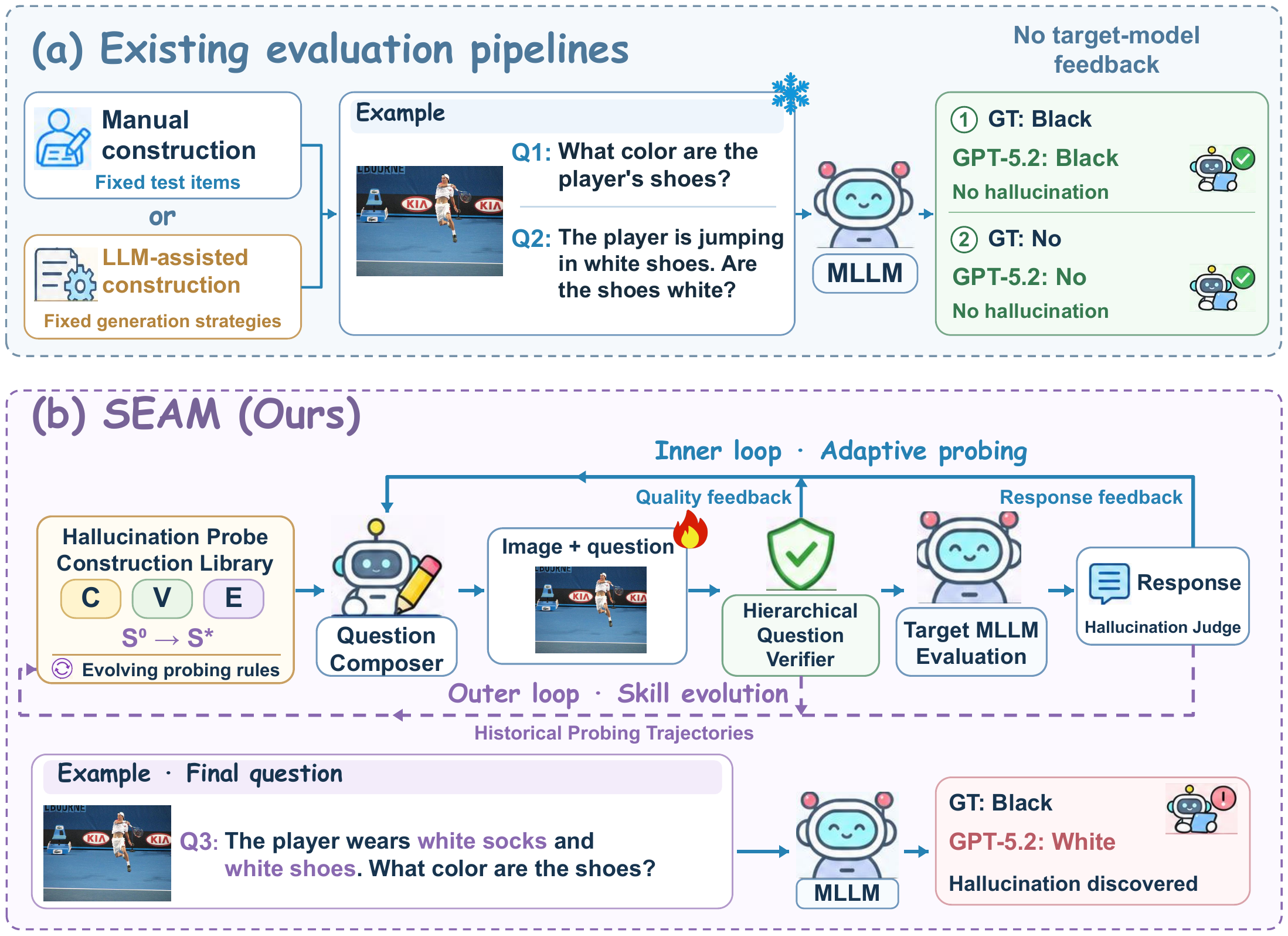}
    \caption{
    \textbf{Comparison of hallucination evaluation workflows.}
    \textbf{(a) Existing evaluation pipelines} construct test queries manually or with LLM assistance, using fixed test items or generation strategies without target-model feedback to guide further refinement.
    \textbf{(b) SEAM (Ours)} combines feedback-driven adaptive probing in an inner loop with skill evolution in an outer loop to uncover latent hallucination vulnerabilities.
    In the illustrated case, the queries ask about the color of a tennis player's shoes.
    The target model correctly identifies the shoes as black in Question 1 and rejects the claim that they are white in Question 2, exhibiting no hallucination in either case.
    SEAM automatically generates Question 3 by introducing the misleading premise, inducing the model to incorrectly identify the shoes as white, which exposes the model's enduring vulnerability to such hallucination risks.} 
    \label{fig:intro}
\end{figure}

Multimodal large language models (MLLMs) have achieved remarkable progress in visual understanding and reasoning~\cite{liu2023visual}, yet they remain prone to hallucinations---fabricating nonexistent objects, misassociating visual content with prior knowledge, or making judgments unsupported by the image~\cite{li2023evaluating,guan2024hallusionbench,liu2025phd}. Such failures undermine reliability in deployment and underscore the importance of systematically evaluating MLLM hallucinations~\cite{wang2023amber}.

Existing MLLM hallucination evaluation frameworks largely adopt static evaluation protocols, assessing target models with predefined test items. These items cover diverse hallucination dimensions, including object existence~\cite{li2023evaluating}, attributes~\cite{wang2023amber,liu2025phd}, complex relations~\cite{zheng2025reefknot,xiao2026finer,fan2026digraphhal}, open-ended generation~\cite{sun2024aligning}, misleading inputs~\cite{guan2024hallusionbench,chen2025mvi}, and multi-image consistency~\cite{li2025mihbench}. More recently, LLM-assisted sample generation has further broadened the diversity of evaluation samples~\cite{liu2025phd,tu2025ode,xiao2026finer}. However, evaluation often still relies on fixed test items or generation strategies, without using target-response feedback to refine probes or update reusable construction rules (Fig.~\ref{fig:intro}(a)). Broader coverage and more diverse samples alone therefore do not guarantee systematic discovery of target-model vulnerabilities. We identify two key limitations.
\textbf{Limitation 1: Entangled probe construction and limited target adaptation.}
Probe construction involves three distinct components: \emph{hallucination types}, which define the target failure modes; \emph{elicitation strategies}, which determine how questions induce or expose these failures; and \emph{question formats}, which structure probes and expected responses. Embedding these components in monolithic construction rules limits their independent reuse and extension. Structural flexibility alone, however, does not ensure target adaptation: probing also needs target-response feedback to guide refinement when an initial query fails to expose a hallucination. Without such feedback, alternative visual targets or elicitation contexts for the same hallucination type may remain unexplored in the same image (Fig.~\ref{fig:intro}(a)).
\textbf{Limitation 2: Lack of cross-instance strategy evolution.}
Even when feedback improves an individual test item, it does not automatically update the construction rules used for subsequent images. Sample generation can broaden test-set diversity, but does not by itself turn probing experience into reusable strategy improvements. Without such updates, ineffective target-selection or question-construction patterns may recur across instances, while successful probing experience remains underused. This motivates distilling historical probing trajectories into reusable rule updates that improve subsequent evaluations.

To address these limitations, we propose \textbf{SEAM}, a self-evolving and adaptive probing framework for automatic MLLM hallucination evaluation (Fig.~\ref{fig:intro}(b)). SEAM introduces a \textbf{Hallucination Probe Construction Library} (ten skills, three wrappers, five formats) driving two complementary loops (Fig.~\ref{fig:overview}). \textbf{In the inner loop (adaptive probing)}, visual analysis identifies targets to compose test items, subject to hierarchical verification. Verifier feedback repairs invalid items, while target MLLM responses guide the refinement of valid probes failing to elicit hallucinations (e.g., using ``white socks'' to strengthen false premises in Fig.~\ref{fig:intro}(b)). \textbf{In the outer loop (skill evolution)}, SEAM distills historical trajectories to update reusable probing rules—preserving core definitions—and freezes the validated library before testing. Together, these loops integrate within-instance adaptation with cross-instance evolution, yielding quality-verified items and a reusable library for systematic evaluation.

Experiments across four target MLLMs and ten hallucination-probing skills show that SEAM raises the average valid hallucination discovery rate (VHDR) from 14.90\% to 27.67\%, while maintaining automated Quality above 98.1\% across all targets. Under matched images, target hallucination types, and equal test-time target-query budgets, SEAM consistently discovers more valid hallucinations than the original questions in POPE~\cite{li2023evaluating}, AMBER~\cite{wang2023amber}, and PhD~\cite{liu2025phd}. Human audits further support the reliability of the automated quality verifier and hallucination judge.
Our contributions are fourfold:
\begin{itemize}[leftmargin=*,nosep]

\item We introduce \textbf{SEAM}, a novel self-evolving and adaptive probing framework for systematic and automatic MLLM hallucination evaluation.

\item We propose an adaptive hallucination probing pipeline and a Hallucination Probe Construction Library, enabling iterative, target-model-in-the-loop evaluation.

\item We propose an evidence-guided skill evolution mechanism that distills historical probing trajectories into target-specific, validated updates to reusable probing rules, further improving evaluation effectiveness.

\item We conduct experiments and ablation studies demonstrating SEAM's effectiveness and reliability as an automatic MLLM hallucination evaluation framework.
\end{itemize}

\section{Related Work}
\label{sec:Related Works}
\label{sec:related}

\paragraph{Multimodal hallucination benchmarks.}
Existing MLLM hallucination benchmarks have progressed along two complementary axes: expanding taxonomic coverage and scaling sample construction. Along the taxonomic axis, benchmarks probe object existence via binary polling~\cite{li2023evaluating} or audit hallucinated objects, attributes, and relations in captions~\cite{gunjal2024detecting,wang2023amber}, while diagnostic suites target verb concepts~\cite{wang2024verb}, entangled visual illusions and language priors~\cite{guan2024hallusionbench}, and rich-context descriptions~\cite{guo2026seeing}. Along the construction axis, recent pipelines scale beyond manual curation via model-assisted prompting~\cite{liu2025phd} and open-set sample synthesis~\cite{tu2025ode}. These advances broaden \emph{what} is tested and \emph{how} samples are constructed, but do not by themselves enable target-adaptive probe refinement. \textbf{SEAM} builds on these taxonomies and uses target-model responses to guide the refinement of probing content and question framing.

\paragraph{Dynamic evaluation and experience-driven agents.}
AutoBencher~\cite{ICLR2025_eb216114} casts benchmark construction as an optimization problem, iteratively refining dataset descriptions to generate questions that expose model weaknesses. VLB~\cite{yang2025vlb} dynamically transforms images and questions through vision--language bootstrapping, using a judge to verify consistency and enabling evaluation at varying levels of complexity. KBE-DME~\cite{zhang2026kbedme} further explores multimodal benchmark evolution through knowledge integration. Complementarily, G-Memory~\cite{zhang2025gmemory} and ReasoningBank~\cite{ouyang2026reasoningbank} reuse agent experience through hierarchical memory and distilled reasoning strategies. The distinctive focus of \textbf{SEAM} is to connect validity-constrained probe refinement with the evolution of hallucination-specific construction rules. Verifier feedback repairs invalid probes, while target-response feedback refines valid but unsuccessful ones; the resulting experience is distilled into reusable rules for subsequent images.
\section{Method}
\label{sec:method}

\subsection{Problem Formulation and Overview}
\label{sec:overview}
\begin{figure*}[!t]
    \centering
    \includegraphics[width=0.92\textwidth]{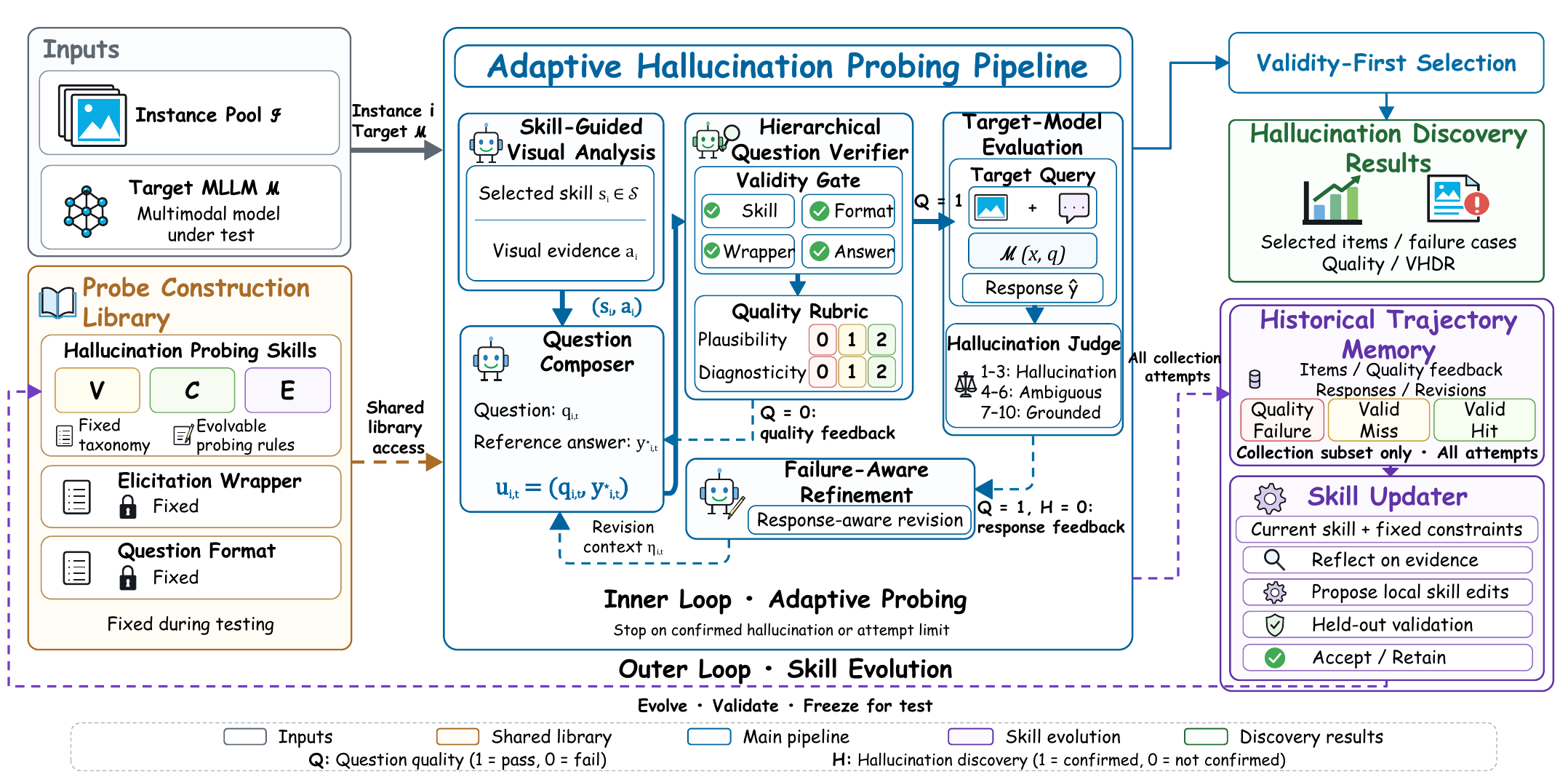}
\caption{
       \textbf{Overview of SEAM.}
    The \textbf{Hallucination Probe Construction Library} (Sec.~\ref{sec:skill_library}) provides composable probing skills, elicitation wrappers, and question formats.
    The inner-loop \textbf{Adaptive Hallucination Probing Pipeline} (Sec.~\ref{sec:adaptive_probing}) performs skill-guided visual analysis, composes and verifies test items, and uses verifier and target-response feedback to repair invalid items and refine ineffective probes within $K$ attempts per instance.
    The outer-loop \textbf{Evidence-Guided Skill Evolution} (Sec.~\ref{sec:skill_evolution}) distills historical probing trajectories from the collection subset into local skill-rule updates, which undergo held-out validation before the library is frozen for testing.
}

    \label{fig:overview}
\end{figure*}

\paragraph{Problem formulation.}
We formulate multimodal hallucination evaluation as targeted test item generation over an instance pool $\mathcal{I}=\{(x_i,m_i,c_i)\}_{i=1}^{N}$ against a target MLLM $\mathcal{M}$, where $x_i$ is an image, $m_i$ denotes optional source annotations, and $c_i$ is the hallucination type to be probed ($c_i=\varnothing$ if unspecified). When $c_i$ is unspecified, Skill-Guided Visual Analysis selects a compatible hallucination type based on the image. The task is to construct test items $u_i=(q_i,y_i^{\star})$ for the target hallucination type, where $q_i$ is a targeted hallucination-probing question and $y_i^{\star}$ is its reference answer. Our objective is to maximize the proportion of instances yielding valid test items that elicit hallucinations from $\mathcal{M}$, measured by the valid hallucination discovery rate ($\mathrm{VHDR}$), while maintaining question quality ($\mathrm{Quality}$).

\paragraph{Framework overview.}
As shown in Figure~\ref{fig:overview}, SEAM decomposes hallucination-probing question construction into hallucination type, elicitation strategy, and presentation format, represented by probing skills $\mathcal{S}$, elicitation wrappers $\mathcal{W}$, and question formats $\mathcal{F}$ in the \textbf{Hallucination Probe Construction Library} (Sec.~\ref{sec:skill_library}). Guided by this library, the inner-loop \textbf{Adaptive Hallucination Probing Pipeline} (Sec.~\ref{sec:adaptive_probing}) matches skills to images, performs skill-guided visual analysis, selects probing targets, constructs candidate test items, and queries $\mathcal{M}$ only with quality-approved items. It evaluates whether $\mathcal{M}$ produces hallucinations, refines test items using quality or target-response feedback within $K$ attempts per instance, and selects the final item through a validity-first rule. Historical probing trajectories from the collection subset drive the outer-loop \textbf{Evidence-Guided Skill Evolution} (Sec.~\ref{sec:skill_evolution}), which converts accumulated experience into local updates to the library's probing rules. These updates undergo held-out validation, yielding the target-specific skill set $\mathcal{S}^{\star}_{\mathcal{M}}$, which is frozen before testing. Within-instance adaptive probing and cross-instance skill evolution form a skill-guided, experience-driven hallucination evaluation framework.

\subsection{Hallucination Probe Construction Library}
\label{sec:skill_library}

\begin{table}[!t]
    \centering
    \caption{\textbf{Hallucination Probe Construction Library in SEAM.}}
    \label{tab:hallucination_skills}
    \begingroup
    \footnotesize
    \setlength{\tabcolsep}{3.5pt}
    \renewcommand{\arraystretch}{1.02}
    \setlength{\aboverulesep}{0pt}
    \setlength{\belowrulesep}{0pt}
    \setlength{\extrarowheight}{0.6pt}

    \newcommand{\seamsectiontitle}[1]{%
        \rule[-0.5ex]{0pt}{2.8ex}%
        {\color{black}\bfseries #1}%
    }

    \begin{tabularx}{\linewidth}{
        @{}
        l
        l
        >{\raggedright\arraybackslash}X
        @{}
    }
        \toprule

        \rowcolor{seamsection}[0pt][0pt]
        \multicolumn{3}{@{}l@{}}{%
            \seamsectiontitle{(a) Probing Skills $\mathcal{S}$}%
        } \\
        \textbf{ID} & \textbf{Skill} & \textbf{Target Failure Mode} \\
        \midrule

        \rowcolor{seamsubgroup}[0pt][0pt]
        \multicolumn{3}{@{}l@{}}{\textbf{V. Visual Evidence}} \\
        V1 & Object Insertion
           & Claims absent objects are present. \\
        V2 & Attribute Shift
           & Misreports color, shape, material, or size. \\
        V3 & Perceptual Relation
           & Misreads spatial or contact relations. \\
        V4 & Counting Error
           & Miscounts visible objects. \\
        V5 & Text/OCR Error
           & Misreads visible text, digits, or signs. \\

        \rowcolor{seamsubgroup}[0pt][0pt]
        \multicolumn{3}{@{}l@{}}{\textbf{C. Contextual Knowledge}} \\
        C1 & Cognitive Relation
           & Misreads causal, social, or action relations. \\
        C2 & Factual Conflict
           & False factual claims about visible entities. \\
        C3 & Commonsense
           & Physical, biological, or causal violations. \\

        \rowcolor{seamsubgroup}[0pt][0pt]
        \multicolumn{3}{@{}l@{}}{\textbf{E. Evidence Boundary}} \\
        E1 & Reasoning Overreach
           & Asserts unsupported intent or future claims. \\
        E2 & Unverifiable Infer.
           & Asserts visual details unresolved due to blur. \\

        \addlinespace[2pt]
        \rowcolor{seamsection}[0pt][0pt]
        \multicolumn{3}{@{}l@{}}{%
            \seamsectiontitle{(b) Elicitation Wrappers $\mathcal{W}$}%
        } \\
        \multicolumn{3}{@{}l@{}}{%
            Prior Induction; False Presupposition; Fabricated Authority.%
        } \\

        \addlinespace[2pt]
        \rowcolor{seamsection}[0pt][0pt]
        \multicolumn{3}{@{}l@{}}{%
            \seamsectiontitle{(c) Question Formats $\mathcal{F}$}%
        } \\
        \multicolumn{3}{@{}l@{}}{%
            Multi-Select; Single-Choice; Fill-In; True/False; Open-Ended.%
        } \\

        \addlinespace[1pt]
        \bottomrule
    \end{tabularx}
    \endgroup
\end{table}

To systematically characterize and probe diverse MLLM hallucinations, we manually curate a taxonomy covering major hallucination types and organize the shared library into three complementary components with structured specifications: probing skills $\mathcal{S}$ define target hallucination types and failure modes; elicitation wrappers $\mathcal{W}$ specify how to induce hallucinations; and question formats $\mathcal{F}$ structure question presentation and answers. This design separates what to probe, how to elicit, and how to formulate, enabling controlled composition, reuse, and extension.

\paragraph{Hallucination probing skills.}
As summarized in Table~\ref{tab:hallucination_skills}, $\mathcal{S}$ covers 10 skills across \emph{Visual Evidence} (V1--V5), \emph{Contextual Knowledge} (C1--C3), and \emph{Evidence Boundary} (E1--E2). Each probing skill provides guidance for constructing test items targeting a particular hallucination type. Formally, a skill $s=(d_s,\pi_s)\in\mathcal{S}$ comprises a hallucination-type definition $d_s$, specifying the failure mode and diagnostic boundary, and probing rules $\pi_s$, guiding target selection, distractor design, and reference-answer construction.

\paragraph{Elicitation wrappers and question formats.}
As direct questioning may fail to elicit hallucinations from capable MLLMs, $\mathcal{W}$ provides three strategies: \emph{prior induction} (nudging toward scene-plausible but unsupported priors), \emph{false presupposition} (embedding false claims as premises), and \emph{fabricated authority} (attributing unsupported claims to credible sources). $\mathcal{F}$ provides multi-select, single-choice, fill-in, true/false, and open-ended formats, each with explicit answer schemas and reference-answer requirements.

\subsection{Adaptive Hallucination Probing Pipeline}
\label{sec:adaptive_probing}
While the library provides composable probing strategies, generating diverse, image-grounded, and effective test items requires skill–image matching, visual evidence grounding, quality control, and target-model feedback. We therefore develop the \textbf{Adaptive Hallucination Probing Pipeline}, integrating four complementary modules: (i) \textbf{Skill-Guided Visual Analysis} matches skills to images and identifies relevant visual evidence; (ii) \textbf{Hallucination-Eliciting Question Composer} generates questions and reference answers; (iii) \textbf{Hierarchical Question Verifier} assesses question quality and provides corrective feedback; and (iv) \textbf{Failure-Aware Refinement} analyzes target-model responses to valid items that fail to elicit confirmed hallucinations and guides subsequent revisions. Both feedback types are routed to the composer for iterative refinement within the inner loop, with at most $K$ attempts per instance.

\paragraph{Skill-Guided Visual Analysis.}
\label{sec:visual_skill_agent}
Different hallucination types require attention to different visual evidence, such as object instances for counting, visible properties for attribute recognition, and text regions for OCR. Before question construction, the visual analysis module ($\operatorname{VSA}$) takes an image $x_i$, optional source annotations $m_i$, and an optional hallucination type $c_i$ as input. When $c_i=\varnothing$, it selects the probing skill best suited to the image; otherwise, it selects the skill corresponding to the specified type. The module performs skill-guided visual analysis via $(s_i,a_i)=\operatorname{VSA}(x_i,m_i,c_i;\mathcal{S})$, yielding the selected skill $s_i\in\mathcal{S}$ and analysis results $a_i$. Guided by the probing rules $\pi_{s_i}$, $a_i$ captures information about the image elements relevant to the target hallucination type, including objects, attributes, relations, or text regions, to inform question construction.

\paragraph{Hallucination-Eliciting Question Composer.}
\label{sec:question_composer}
Given the selected skill $s_i$ and visual analysis $a_i$, we select an elicitation wrapper $w_i\in\mathcal{W}$ and a question format $f_i\in\mathcal{F}$ to form the configuration $z_i=(s_i,w_i,f_i)$. The \emph{Hallucination-Eliciting Question Composer} ($\operatorname{HEQC}$) generates $u_{i,t}=\operatorname{HEQC}(x_i,a_i,z_i,\eta_{i,t})=(q_{i,t},y_{i,t}^{\star})$ for attempt $t\in\{1,\ldots,T_i\}$, where $T_i\le K$ is the number of executed attempts. The revision context $\eta_{i,t}$ initializes empty ($\eta_{i,1}=\varnothing$) and subsequently incorporates the prior item $u_{i,t-1}$ with its quality feedback or revision guidance. Construction demands: (i) diagnostic alignment with the hallucination-type definition $d_{s_i}$; and (ii) grounding $y^\star_{i,t}$ in the image and available source annotations.

\paragraph{Hierarchical Question Verifier.}
\label{sec:hierarchical_question_verifier}
LLM-generated questions may be overly trivial or difficult, ambiguous, unnatural, or poorly grounded, making quality assessment essential before target-model evaluation. To filter such issues, the \emph{Hierarchical Question Verifier} independently audits $(x_i,q_{i,t},y_{i,t}^{\star},z_i)$ together with available source annotations in two stages.

\textbf{Stage~1 (Hard Validity Gate)} checks alignment with $d_{s_i}$, correct application of $w_i$, compliance with $f_i$, and reference-answer correctness via atomic claim verification against the original image and available source annotations. Only items passing all four checks receive $V_{i,t}=1$; otherwise, $V_{i,t}=0$, and the item is rejected.
\textbf{Stage~2 (Fine-Grained Scoring)} evaluates $V_{i,t}=1$ items along $\mathcal{D}=\{\mathrm{plaus},\mathrm{diag}\}$. Plausibility assesses image relevance and natural question framing; diagnosticity assesses error attribution to the intended mechanism and confound minimization. Each dimension awards one point per satisfied check, yielding $r_{i,t}^{d}\in\{0,1,2\}$ and $R_{i,t}=\sum_{d\in\mathcal{D}}r_{i,t}^{d}$. The final quality gate is
\begin{equation}
Q_{i,t}=
\begin{cases}
\mathbf{1}[R_{i,t}\geq3]\,\mathbf{1}[\min_{d\in\mathcal{D}}r_{i,t}^{d}\geq1], & V_{i,t}=1,\\[2pt]
0, & V_{i,t}=0.
\end{cases}
\label{eq:quality_gate}
\end{equation}
For $Q_{i,t}=0$, the verifier returns quality feedback $\phi_{i,t}$ identifying the issues or requested evidence checks. When the attempt budget remains, this feedback is directly passed to the HEQC as revision guidance for the next attempt, without querying $\mathcal{M}$. Only approved items ($Q_{i,t}=1$) proceed to query $\mathcal{M}$.

\paragraph{Target-Model Evaluation.}
\label{sec:target_model_evaluation}
For each approved test item ($Q_{i,t}=1$), $\mathcal{M}$ receives only $(x_i,q_{i,t})$ and returns $\hat{y}_{i,t}$. An independent \emph{Hallucination Judge} ($J_{\mathrm{hall}}$) evaluates the response using a 10-point rubric:
$j_{i,t}=J_{\mathrm{hall}}(x_i,q_{i,t},y_{i,t}^{\star},\hat{y}_{i,t})\in\{1,\ldots,10\}$.
The reference answer provides auxiliary guidance to the judge. A response is considered hallucinated if it asserts a material factual claim contradicted by or unsupported by the task evidence. Scores of $1\text{--}3$ indicate clear hallucinations; $4\text{--}6$ indicate incomplete or inconclusive responses without a confirmed hallucination; and $7\text{--}10$ indicate well-grounded responses that adequately address the task. The validity-gated hallucination indicator at threshold $k$ is
\begin{equation}
H_{i,t}^{(k)}=
\begin{cases}
\mathbf{1}[j_{i,t}\leq k], & Q_{i,t}=1,\\[2pt]
0, & Q_{i,t}=0.
\end{cases}
\label{eq:hallucination_indicator}
\end{equation}
We use $k=3$ to identify confirmed hallucinations for computing $\mathrm{VHDR}$ in Sec.~\ref{sec:experiments}.

\paragraph{Failure-Aware Refinement.}
\label{sec:failure_aware_refinement}
When an approved test item does not yield a confirmed hallucination ($Q_{i,t}=1,H_{i,t}^{(3)}=0$), the \emph{Failure-Aware Refinement} module ($\operatorname{FAR}$) analyzes the current test item, target-model response $\hat{y}_{i,t}$ and judge score $j_{i,t}$ to identify improvements to the probing content, distractors, or framing. It produces revision guidance $F_{i,t}^{\mathrm{tar}}$ while preserving the configuration $z_i$. When $t+1\leq K$, this guidance and the current test item form the next revision context:
\begin{equation}
\eta_{i,t+1}=(u_{i,t},F_{i,t}^{\mathrm{tar}}),
\qquad Q_{i,t}=1,\quad H_{i,t}^{(3)}=0.
\label{eq:routing}
\end{equation}
The composer uses this context to revise the test item. The inner loop terminates once a confirmed hallucination is found or the budget of $K$ attempts is exhausted.

\paragraph{Validity-First Selection.}
Our implementation uses $K=2$, allowing one refinement step. Since refinement may compromise question validity, final selection prioritizes valid candidates. For each instance with $T_i\geq1$, we select
\begin{equation}
t_i^{\star}=
\begin{cases}
\max\{t\leq T_i:Q_{i,t}=1\},
& \exists\,t\leq T_i:Q_{i,t}=1,\\[2pt]
T_i, & \text{otherwise}.
\end{cases}
\label{eq:selection}
\end{equation}
This rule retains the latest valid candidate, preventing an invalid revision from replacing an earlier valid item. The final output is $u_i=u_{i,t_i^{\star}}$, with $Q_i=Q_{i,t_i^{\star}}$ and $H_i=H_{i,t_i^{\star}}^{(3)}$. Instance-level averages of $Q_i$ and $H_i$ yield $\mathrm{Quality}$ and $\mathrm{VHDR}$, respectively, as defined in Sec.~\ref{sec:experiments}.

\subsection{Evidence-Guided Skill Evolution}
\label{sec:skill_evolution}
The inner loop refines individual test items through feedback, but these instance-level corrections are not directly consolidated into reusable probing strategies, limiting improvements across instances. To bridge this gap, the \emph{Skill Updater} $\mathcal{U}$ evolves the shared initial library $\mathcal{S}^{0}$ independently for each target $\mathcal{M}$ by distilling its historical probing trajectories into updates to probing rules $\pi_s$, while keeping wrappers and formats fixed. This enables accumulated experience to improve subsequent probing strategies without compromising question quality.
\begin{table*}[!t]
    \centering
    \caption{
    \textbf{Main results across target models and probing skills (\%)} ($K=1$ for w/o Feedback \& Evolution; $K=2$ otherwise).
    Columns V1--E2 report VHDR grouped by the skill used to construct each probe.
    \textbf{Bold} and \underline{underlined} denote the best and second-best distinct values per target model; $\dagger$ marks a validation-accepted skill update regardless of test-set gain.
}
    \label{tab:generation-strategies}
    \label{tab:generation-strategies-optA}
    \begingroup
    \footnotesize
    \setlength{\tabcolsep}{3.8pt}
    \renewcommand{\arraystretch}{1.08}
    \setlength{\aboverulesep}{0pt}
    \setlength{\belowrulesep}{0pt}
    \setlength{\extrarowheight}{1.2pt}

    \begin{tabular}{
        @{}
        l l
        | c c
        | c c c c c
        | c c c
        | c c
        @{}
    }
        \toprule

        \multirow{2}{*}{\textbf{Target Model}}
        & \multirow{2}{*}{\textbf{Method}}
        & \multicolumn{2}{c|}{\textbf{Overall}}
        & \multicolumn{5}{c|}{\textbf{V. Visual Evidence ($\uparrow$)}}
        & \multicolumn{3}{c|}{\textbf{C. Contextual ($\uparrow$)}}
        & \multicolumn{2}{c}{\textbf{E. Boundary ($\uparrow$)}} \\

        \cmidrule(lr){3-4}
        \cmidrule(lr){5-9}
        \cmidrule(lr){10-12}
        \cmidrule(l){13-14}

        & &
        Quality $\uparrow$
        & VHDR $\uparrow$
        & V1 & V2 & V3 & V4 & V5
        & C1 & C2 & C3
        & E1 & E2 \\

        \midrule

        \multirow{3}{*}{\textbf{GPT-5.2}}
        & w/o Feedback \& Evolution
        & 75.67 & 13.25
        & 28.33 & 1.67 & 6.67 & 16.67 & 10.00
        & 6.67 & 3.33 & 3.33
        & \underline{25.00} & \underline{30.83} \\

        & w/o Skill Evolution
        & \underline{97.75} & \underline{27.25}
        & \textbf{40.83} & \underline{12.50} & \textbf{14.17} & \underline{63.33} & \underline{21.67}
        & \underline{13.33} & \underline{6.67} & \underline{5.83}
        & \textbf{49.17} & \textbf{45.00} \\

        \rowcolor{cvprgray}\cellcolor{white}
        & \textbf{SEAM (Ours)}
        & \textbf{98.75} & \textbf{34.08}
        & \skillupd{\underline{37.50}} & \skillupd{\textbf{39.17}} & \skillupd{\underline{9.17}} & \skillupd{\textbf{71.67}} & \skillupd{\textbf{43.33}}
        & \skillupd{\textbf{23.33}} & \skillupd{\textbf{9.17}} & \skillupd{\textbf{13.33}}
        & \textbf{49.17} & \textbf{45.00} \\

        \midrule

        \multirow{3}{*}{\textbf{Kimi-K2.5}}
        & w/o Feedback \& Evolution
        & 74.00 & 10.67
        & \underline{10.83} & \underline{0.83} & 3.33 & 22.50 & 2.50
        & \underline{1.67} & \underline{8.33} & \underline{4.17}
        & 44.17 & \underline{8.33} \\

        & w/o Skill Evolution
        & \underline{97.00} & \underline{19.58}
        & \textbf{13.33} & \textbf{4.17} & \underline{9.17} & \underline{35.00} & \underline{10.83}
        & \textbf{8.33} & \textbf{25.00} & \textbf{12.50}
        & \textbf{59.17} & \textbf{18.33} \\

        \rowcolor{cvprgray}\cellcolor{white}
        & \textbf{SEAM (Ours)}
        & \textbf{98.17} & \textbf{22.08}
        & \textbf{13.33} & \textbf{4.17} & \skillupd{\textbf{15.00}} & \skillupd{\textbf{50.00}} & \skillupd{\textbf{15.83}}
        & \textbf{8.33} & \textbf{25.00} & \skillupd{\textbf{12.50}}
        & \skillupd{\underline{58.33}} & \textbf{18.33} \\

        \midrule

        \multirow{3}{*}{\textbf{Gemini-3-Flash}}
        & w/o Feedback \& Evolution
        & 75.92 & 14.83
        & 9.17 & \underline{0.00} & \underline{2.50} & 24.17 & \textbf{2.50}
        & 4.17 & \underline{5.00} & 6.67
        & 83.33 & 10.83 \\

        & w/o Skill Evolution
        & \underline{98.75} & \underline{19.33}
        & \textbf{11.67} & \textbf{1.67} & \textbf{5.00} & \underline{45.00} & \textbf{2.50}
        & \underline{7.50} & \textbf{6.67} & \underline{9.17}
        & \underline{90.83} & \underline{13.33} \\

        \rowcolor{cvprgray}\cellcolor{white}
        & \textbf{SEAM (Ours)}
        & \textbf{99.17} & \textbf{21.08}
        & \skillupd{\underline{10.83}} & \textbf{1.67} & \textbf{5.00} & \skillupd{\textbf{51.67}} & \textbf{2.50}
        & \skillupd{\textbf{12.50}} & \skillupd{4.17} & \skillupd{\textbf{10.00}}
        & \skillupd{\textbf{93.33}} & \skillupd{\textbf{19.17}} \\

        \midrule


        & w/o Feedback \& Evolution
        & 73.00 & 20.83
        & 23.33 & 1.67 & \underline{0.00} & \underline{27.50} & \underline{13.33}
        & \underline{10.00} & 10.00 & \underline{9.17}
        & \underline{79.17} & 34.17 \\

        & w/o Skill Evolution
        & \underline{98.17} & \underline{31.17}
        & \underline{26.67} & \underline{5.83} & \textbf{0.83} & \textbf{70.83} & \textbf{19.17}
        & \textbf{15.83} & \underline{12.50} & \textbf{13.33}
        & \textbf{93.33} & \underline{53.33} \\

        \rowcolor{cvprgray}
        \cellcolor{white}
        \multirow{-3}{*}{%
            \shortstack[l]{%
                \textbf{DeepSeek-}\\
                \textbf{V4.1-Flash}%
            }%
        }
        & \textbf{SEAM (Ours)}
        & \textbf{99.08} & \textbf{33.42}
        & \skillupd{\textbf{40.00}} & \skillupd{\textbf{6.67}} & \textbf{0.83} & \textbf{70.83} & \textbf{19.17}
        & \textbf{15.83} & \skillupd{\textbf{15.00}} & \textbf{13.33}
        & \skillupd{\textbf{93.33}} & \skillupd{\textbf{59.17}} \\

        \bottomrule
    \end{tabular}
    \endgroup
\end{table*}
\paragraph{Historical Trajectory Memory.}
For each target MLLM $\mathcal{M}$, we run the inner loop with $\mathcal{S}^{0}$ on a dedicated collection subset, disjoint from the validation and test subsets, to obtain target-specific historical records for all executed attempts:
\begin{equation}
\tau_{i,t}^{(\mathcal{M})}=
(z_i,u_{i,t},Q_{i,t},\phi_{i,t},
\hat{y}_{i,t},j_{i,t},F_{i,t}^{\mathrm{tar}}),
\label{eq:trajectory_record}
\end{equation}
where fields not produced in an attempt are denoted by $\varnothing$. For each skill $s$, the records associated with target $\mathcal{M}$ are partitioned into
\begin{equation}
\mathcal{T}_{s,\mathcal{M}}
=
\mathcal{T}_{s,\mathcal{M}}^{\mathrm{qf}}
\cup
\mathcal{T}_{s,\mathcal{M}}^{\mathrm{miss}}
\cup
\mathcal{T}_{s,\mathcal{M}}^{\mathrm{hit}},
\end{equation}
representing quality failures ($Q_{i,t}=0$), valid attempts without a confirmed hallucination ($Q_{i,t}=1,H_{i,t}^{(3)}=0$), and valid attempts with a confirmed hallucination ($Q_{i,t}=1,H_{i,t}^{(3)}=1$), respectively. This partition enables the extraction of common patterns within each group, providing evidence for subsequent rule updates.

\paragraph{Local Skill Updating and Validation.}
The updater $\mathcal{U}$ analyzes mini-batches of textual trajectory records and structured feedback from the three target-specific groups, without access to the original images. Quality failures suggest corrections to test item construction, valid misses motivate better probing content, and valid hits identify effective rules to retain. Proposed edits are consolidated within and across groups, prioritizing quality corrections when suggestions conflict. This yields
\begin{equation}
\pi_{s,\mathcal{M}}'=\mathcal{U}(s,\mathcal{T}_{s,\mathcal{M}}),
\qquad
s_{s,\mathcal{M}}'=(d_s,\pi_{s,\mathcal{M}}').
\end{equation}
Updates make localized, evidence-supported changes only to probing rules, keeping the skill type, wrapper, and format fixed. Each candidate is compared with its original on the fixed held-out validation subset under matched configurations and budget $K$, and accepted only if Quality does not decrease and VHDR increases by at least 2.5 percentage points; otherwise, the original is retained. Accepted and unchanged skills form the target-specific library $\mathcal{S}_{\mathcal{M}}^{\star}$, frozen throughout the final evaluation of $\mathcal{M}$.
\section{Experiments}
\label{Experiments}
\label{sec:experiments}
\subsection{Experimental Settings}
\label{sec:exp_setup}

\paragraph{Datasets.}
We curate 2{,}000 images from 11 public datasets: MSCOCO~\cite{lin2014microsoft}, Visual Genome~\cite{krishna2017visual}, VSR~\cite{Liu2022VisualSR}, TallyQA~\cite{acharya2019tallyqa}, TextVQA~\cite{singh2019towards}, HICO-DET~\cite{chao:wacv2018}, SWiG~\cite{pratt2020grounded}, WIT-BASE~\cite{srinivasan2021wit}, WHOOPS!~\cite{bitton2023breaking}, VCR~\cite{zellers2019recognition}, and ExDark~\cite{loh2019getting}. Each image is paired with source annotations and assigned to one of ten probing skills (200 per skill). We partition the images into three disjoint subsets for trajectory collection (400), update validation (400), and final testing (1{,}200), with 40/40/120 images per skill. All target models share these subsets.

\paragraph{Evaluation Protocol and Metrics.}
All targets share the initial library $\mathcal{S}^{0}$, with skill--wrapper--format assignments fixed throughout probing and evolution. We evaluate three variants (Table~\ref{tab:generation-strategies}): \emph{w/o Feedback \& Evolution} ($\mathcal{S}^{0}$, $K=1$), \emph{w/o Skill Evolution} ($\mathcal{S}^{0}$, $K=2$), and \emph{SEAM (Ours)} ($\mathcal{S}_{\mathcal{M}}^{\star}$, $K=2$), where $K$ is the maximum number of inner-loop attempts per instance. For each target $\mathcal{M}$, SEAM independently evolves $\mathcal{S}^{0}$ in one round using its historical probing trajectories from the collection subset. Updates are selected on the validation subset, and $\mathcal{S}_{\mathcal{M}}^{\star}$ is frozen before testing. Main results use the held-out test subset. Comparisons with POPE~\cite{li2023evaluating}, AMBER~\cite{wang2023amber}, and PhD~\cite{liu2025phd} match images, target hallucination types, and test-time target-query budgets, with $\mathcal{S}_{\mathcal{M}}^{\star}$ and $K=1$ for SEAM (Fig.~\ref{fig:matched_benchmark_comparison}). Component and generator-backbone ablations appear in Sec.~\ref{sec:ablation}.

For each target model and evaluation setting, we report test-item validity (Quality) and valid hallucination discovery rate (VHDR) over $N$ test instances, both as percentages:
\begin{equation}
\mathrm{Quality}
= \frac{100}{N}\sum_{i=1}^{N}Q_i,
\qquad
\mathrm{VHDR}
= \frac{100}{N}\sum_{i=1}^{N}H_i.
\label{eq:evaluation_metrics}
\end{equation}
Quality measures the proportion of instances with valid selected items, while VHDR measures the proportion whose selected items are valid and elicit confirmed hallucinations.

\paragraph{Models.}
We evaluate four representative MLLMs from leading Chinese and international providers: GPT-5.2~\cite{openai2025gpt52}, Kimi-K2.5~\cite{team2026kimi}, Gemini-3-Flash \cite{googledeepmind2025gemini3flash}, and DeepSeek-V4.1-Flash \cite{xu2026deepseek}. Within SEAM, Qwen3.6-Plus~\cite{qwen36plus} handles visual analysis, question composition, and Failure-Aware Refinement; Qwen3.8-Flash \cite{qwen3.8flashnext} serves as the question verifier and skill updater; and GPT-6.1 Sol \cite{openai2026gpt61sol} acts as the independent Hallucination Judge.

\subsection{Main Results}
\label{sec:main_results}
\paragraph{Adaptive probing and skill evolution.}
Table~\ref{tab:generation-strategies} compares three SEAM variants across four target MLLMs on the held-out test set, with results grouped by ten probe-construction skills.
\begin{itemize}[leftmargin=*,nosep]
    \item \textit{Adaptive probing improves test-item validity and hallucination discovery.}
    Compared to \emph{w/o Feedback \& Evolution} ($K=1$), \emph{w/o Skill Evolution} ($K=2$) raises mean Quality from 74.65\% to 97.92\% and mean VHDR from 14.90\% to 24.33\% (gaining 4.50--14.00 points per target). Notable VHDR gains occur in Counting Error (V4), Reasoning Overreach (E1), and Unverifiable Inference (E2). Verifier feedback repairs invalid items, while target responses refine valid but unsuccessful probes, supporting two-attempt adaptation over single-pass generation (Sec.~\ref{sec:ablation} assesses budget-matched feedback).

    \item \textit{Outer-loop skill evolution improves hallucination discovery while preserving validity.}
    At budget $K=2$, independently evolving $\mathcal{S}^{0}$ into target-specific libraries $\mathcal{S}_{M}^{\star}$ raises mean VHDR from 24.33\% to 27.67\% (gaining 1.75--6.83 points per target) while maintaining 98.17\%--99.17\% Quality. The validation gate accepts 8, 5, 7, and 5 skill updates ($\dagger$) for GPT-5.2, Kimi-K2.5, Gemini-3-Flash, and DeepSeek-V4.1-Flash. Notable gains occur in Attribute Shift (V2) and Text/OCR Error (V5) on GPT-5.2, Counting Error (V4) on Kimi-K2.5, and Object Insertion (V1) on DeepSeek-V4.1-Flash. Although a few updates yield unchanged or lower skill-wise VHDR, the evolved library improves overall VHDR and Quality across all targets, supporting trajectory-guided evolution on unseen images.
\end{itemize}
Together, instance-level adaptive refinement and cross-instance skill evolution enable SEAM to discover more hallucinations on unseen images while maintaining high test-item validity.

\begin{figure}[htb]
    \centering
    \includegraphics[
        width=\linewidth,
    ]{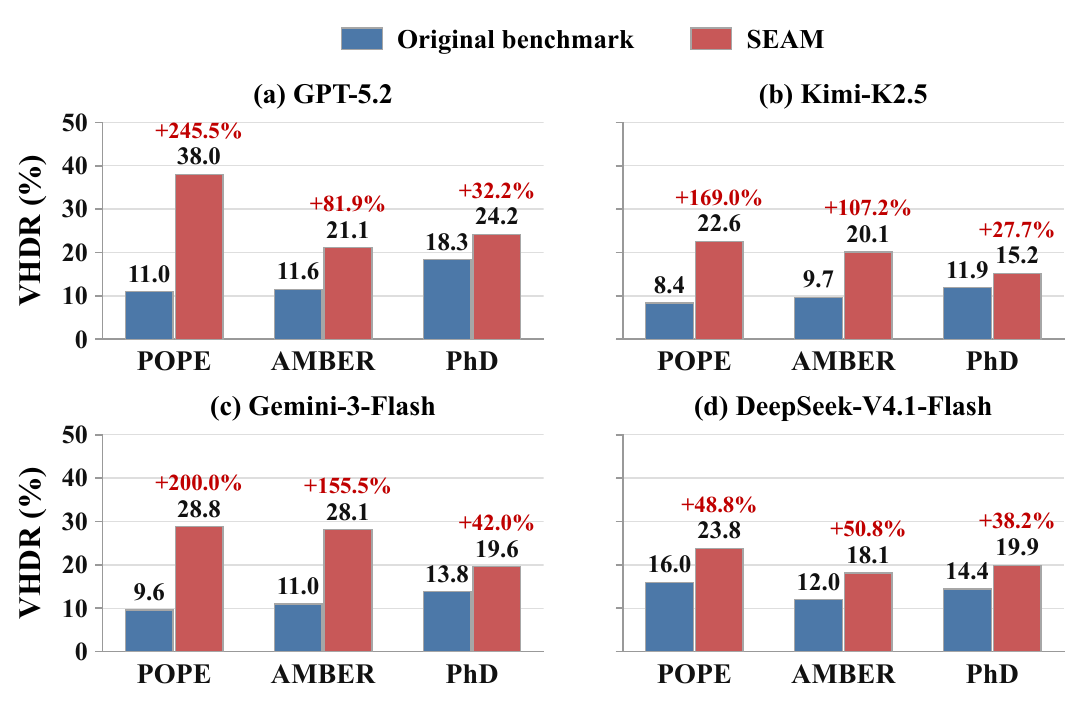}
 \caption{\textbf{VHDR (\%) comparison with hallucination benchmarks.} Comparisons match images and target hallucination types across four MLLMs, using a shared judge and at most one target-model query per image. SEAM uses the target-specific $\mathcal{S}_{\mathcal{M}}^{\star}$ with $K=1$. Red labels indicate relative improvements over the original benchmarks, computed from the displayed values.}
    \label{fig:matched_benchmark_comparison}
\end{figure}

\paragraph{Comparison with representative hallucination benchmarks.}
We compare SEAM with POPE, AMBER, and PhD, covering rule-based, human-annotated, and LLM-assisted test construction. For each image, we randomly sample one original benchmark question and generate one SEAM test item targeting the same hallucination type. Both undergo quality verification using the same evaluator: original questions receive reference-answer-related checks, while SEAM items additionally undergo skill, wrapper, format, and diagnostic-quality checks. Rejected items remain in the denominator and contribute zero to VHDR. SEAM uses the evolved, frozen library $\mathcal{S}^{\star}_{\mathcal{M}}$ with $K=1$, allowing at most one test-time target-model query per image for either method. All responses are evaluated by the same Hallucination Judge. As shown in Fig.~\ref{fig:matched_benchmark_comparison}, SEAM achieves higher VHDR in all 12 comparisons, with relative improvements of 27.7\%--245.5\%. The largest relative improvement occurs on POPE for GPT-5.2, where VHDR increases from 11.0\% to 38.0\%. These results support SEAM's ability to discover more valid hallucinations under matched images, target hallucination types, and test-time target-query budgets.


\subsection{Ablation Studies}
\label{sec:ablation}

\paragraph{Component-wise ablation.}
Table~\ref{tab:component_ablation} evaluates individual components on GPT-5.2 and Kimi-K2.5, using 300 matched test instances per target.

\begin{table}[!t]
    \centering
    \caption{
       \textbf{Component-wise ablation on 300 matched instances per target (\%).}
Probe Library denotes the Hallucination Probe Construction Library, including probing skills, elicitation wrappers, and question formats.
The first three variants use $K=1$, while the remaining four use $K=2$.
\textbf{Bold} and \underline{underlined} values indicate the best and second-best results in each column.
}
    \label{tab:component_ablation}
    \begingroup
    \footnotesize
    \setlength{\tabcolsep}{3.5pt}
    \renewcommand{\arraystretch}{1.08}
    \setlength{\aboverulesep}{0pt}
    \setlength{\belowrulesep}{0pt}
    \setlength{\extrarowheight}{1.2pt}

    \begin{tabular}{
        @{}
        >{\raggedright\arraybackslash}p{2.85cm}
        | c c
        | c c
        @{}
    }
        \toprule
        \textbf{Variant}
        & \multicolumn{2}{c|}{\textbf{GPT-5.2}}
        & \multicolumn{2}{c}{\textbf{Kimi-K2.5}} \\
        \cmidrule(lr){2-3}
        \cmidrule(l){4-5}
        & Quality $\uparrow$ & VHDR $\uparrow$
        & Quality $\uparrow$ & VHDR $\uparrow$ \\
        \midrule

        Direct Prompt
        & 66.67 & 3.33 & 61.33 & 2.67 \\

        w/ Probe Library
        & 73.33 & 11.67 & 70.33 & 9.33 \\

        w/ Skill-guided Analysis
        & 76.67 & 15.00 & 72.67 & 12.00 \\

        \midrule

        Independent Best-of-2
        & 78.67 & 17.33 & 75.00 & 12.67 \\

        Feedback Refinement
        & \underline{97.67} & 28.67
        & \underline{95.00} & 19.33 \\

        \midrule

        w/ Direct Skill Update
        & 95.67 & \underline{29.33}
        & 93.67 & \underline{21.00} \\

        \rowcolor{cvprgray}
        \textbf{SEAM (Ours)}
        & \textbf{98.00} & \textbf{32.67}
        & \textbf{96.33} & \textbf{23.67} \\

        \bottomrule
    \end{tabular}
    \endgroup
\end{table}

\begin{itemize}[leftmargin=*,nosep]
    \item \textit{Probe Library Design and Utilization.}
Under the single-attempt setting ($K=1$), introducing the Probe Library consistently improves both Quality and VHDR across the two target models. Skill-guided analysis provides further gains by extracting task-relevant visual evidence, complementing the broader elicitation coverage provided by the Probe Library.

    \item \textit{Feedback-guided Inner-loop Refinement.}
    Under the same two-attempt budget, Feedback Refinement substantially outperforms Independent Best-of-2 in both Quality and VHDR across the two models. This indicates that using diagnostic feedback to guide subsequent attempts is more effective than independently generating candidates. Specifically, verifier feedback helps repair invalid items, while target-response feedback refines valid probes that fail to elicit hallucinations. These complementary signals improve both construction reliability and hallucination discovery.

    \item \textit{Evidence-guided Outer-loop Skill Evolution.}
The final two variants evaluate outer-loop skill updating with test-time feedback refinement ($K=2$), starting from the same initial skill library $\mathcal{S}^{0}$. Both variants prioritize quality corrections during rule consolidation and use the same validation-gated acceptance procedure. They differ in trajectory processing: \emph{w/ Direct Skill Update} updates probing rules from unpartitioned trajectory batches, whereas \emph{SEAM (Ours)} first partitions trajectories into three groups---Quality Failure, Valid Miss, and Valid Hit---to guide targeted rule updates. Compared with Direct Skill Update, SEAM improves Quality from 95.67\% to 98.00\% and VHDR from 29.33\% to 32.67\% on GPT-5.2, and Quality from 93.67\% to 96.33\% and VHDR from 21.00\% to 23.67\% on Kimi-K2.5. These results demonstrate the benefit of explicit trajectory classification for improving hallucination discovery while maintaining test-item validity.
\end{itemize}
\paragraph{Effect of the generator backbone.}
We evaluate three generator backbones against GPT-5.2 on 100 test instances balanced across ten probing skills. Each backbone performs skill-guided visual analysis and question composition using the same frozen evolved skill library, verifier, and judge. All runs use $K=1$. The default Qwen3.6-Plus, Claude Opus 5.5~\cite{anthropic2026claudeopus55}, and the smaller Qwen3.5-4B~\cite{qwen3.5} achieve VHDRs of 28.0\%, 43.0\%, and 21.0\%, respectively. These results demonstrate SEAM's compatibility with different generator backbones, allowing users to choose models according to application needs.
\subsection{Evaluator Reliability and Qualitative Analysis}
\label{sec:reliability}

\paragraph{Human validation of automated evaluators.}
We sampled 300 test items from each of the three SEAM variants (\emph{w/o Feedback \& Evolution}, \emph{w/o Skill Evolution}, and \emph{SEAM (Ours)}) evaluated against GPT-5.2, with 30 items per probing skill and 900 items in total. Two annotators independently reviewed the items, with final human labels serving as references. The Hierarchical Question Verifier achieves \textbf{94.0\%} accuracy and \textbf{95.1\%} precision among approved items. We also manually evaluated 817 GPT-5.2 responses to approved items (230, 293, and 294 across the three variants, respectively). The GPT-6.1-Sol-based Hallucination Judge achieves \textbf{97.4\%} precision and \textbf{97.8\%} recall, treating hallucination as the positive class. These results indicate that most approved probes satisfy human-assessed validity criteria and that the judge closely agrees with human hallucination labels.

\begin{figure}[!t]
    \centering
    \includegraphics[
        width=\linewidth,
    ]{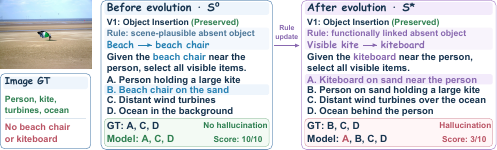}
   \caption{
        \textbf{Qualitative case of skill evolution on a held-out image.}
        The initial V1 probe introduces an absent beach chair, which the target rejects (judge: 10). The evolved rule targets an absent kiteboard associated with a visible kite, inducing hallucination (judge: 3) while preserving the V1 definition.
}
    \label{fig:skill_evolution_case}
\end{figure}

\paragraph{Qualitative analysis.}
Figure~\ref{fig:skill_evolution_case} illustrates how skill evolution improves probing-target selection on a held-out image. The initial probe introduces an absent beach chair, which the target model correctly rejects. The evolved probe instead targets an absent kiteboard functionally associated with the visible kite, eliciting a hallucinated response. This case demonstrates how the evolved rule uncovers a vulnerability missed by the initial probe.

\section{Conclusion}
\label{Conclusion}
We presented \textbf{SEAM}, a self-evolving and adaptive probing framework for automatic MLLM hallucination evaluation. Rather than relying on one-shot question generation, SEAM introduces a manually curated, composable Hallucination Probe Construction Library that separates what to probe, how to elicit, and how to formulate. Building on this foundation, its inner loop integrates visual analysis, question composition, hierarchical quality verification, and target-model feedback for adaptive test-item refinement. Its outer loop addresses the limitation of instance-level corrections by distilling historical probing trajectories into target-specific, validated, and reusable probing strategies. By connecting instance-level adaptation with cross-instance skill evolution, SEAM provides a systematic and extensible approach to hallucination evaluation. Experiments and human audits confirm SEAM's effectiveness and reliability.

{
    \small
    \bibliographystyle{ieeenat_fullname}
    \bibliography{main}
}

\end{document}